\documentclass[11pt,a4paper]{article}
\usepackage[T1]{fontenc}
\usepackage[utf8]{inputenc}
\usepackage{lmodern}
\usepackage{geometry}
\usepackage{microtype}
\usepackage{graphicx}
\usepackage{booktabs}
\usepackage{tabularx}
\usepackage{longtable}
\usepackage{array}
\usepackage{ragged2e}
\usepackage{pdflscape}
\usepackage{enumitem}
\usepackage{xurl}
\usepackage[hidelinks]{hyperref}
\hypersetup{
  pdftitle={The Half-Lives of Generative-AI Evidence: A 40-Record Audit, a Claim-Currency Framework, and a Reflexive Case of Frontier-Model-Assisted Research},
  pdfauthor={Carlo Iacono},
  pdfsubject={Generative-AI evidence currency and frontier-model-assisted research creation},
  pdfkeywords={generative AI, large language models, evidence currency, reproducibility, meta-research, AI-assisted research}
}
\renewcommand{\arraystretch}{1.15}
\newcolumntype{Y}{>{\RaggedRight\arraybackslash}X}

\title{\textbf{The Half-Lives of Generative-AI Evidence}\\[0.45em]
\large A 40-record audit, a claim-currency framework, and a reflexive case of frontier-model-assisted research}
\author{Carlo Iacono\\
\small Charles Sturt University, Australia\\
\small ORCID: \href{https://orcid.org/0009-0003-1668-7396}{0009-0003-1668-7396}}
\date{\small Data cutoff: 17 July 2026 \\ Submission-stage frontier note: 27 July 2026 \\ Article type: purposive empirical audit, reporting framework and reflexive production case \\ Included records: 40}

\begin{document}
\maketitle
\begin{abstract}
Generative-AI evaluations can become historical before publication, yet calendar age does not affect every conclusion equally. This paper has two linked purposes. First, it audits a maximum-variation purposive corpus of 40 empirical records appearing between 18 July 2025 and 17 July 2026. The audit coded publication route, execution timing, model identity, age of the newest named generation or immutable snapshot, same-family supersession and refresh behaviour. At appearance, the newest named model was a median 281 days old (middle 50\%: 75--478; range: 11--939). Median age was 395 days for 25 journal articles, 56 days for 14 preprints and 49 days for one laboratory report. Thirty-five records included a superseded family, seven supplied a precise dated identifier, three clearly refreshed model evidence, and one added a late sensitivity test. All 40 included an OpenAI system, a feature of this corpus rather than a prevalence estimate. The paper distinguishes model age from claim currency and proposes six reporting practices. Second, it treats its own two-day production process as a reflexive case of frontier-model-assisted research creation. GPT-5.6 Sol Pro in ChatGPT supported candidate discovery, source reconciliation, calculations, drafting and critique; the author checked sources, made all substantive decisions and accepts responsibility. This is a proof-of-practice, not a controlled estimate of productivity or quality. By applying its own Model Facts and model-currency statement, the paper shows how rapid AI-assisted research can be made inspectable without treating model output as independent validation. The title uses half-lives metaphorically; no universal decay rate is estimated.
\end{abstract}
\noindent\textbf{Keywords:} generative artificial intelligence; large language models; evidence currency; reproducibility; scholarly communication; meta-research; AI-assisted research
\section{Introduction}
One record in this corpus was accepted three days before publication. Under the audit's coding rule, its newest tested model was already about 287 days old. The paper had moved quickly by journal standards; the systems around it had moved faster. Models can be replaced, rerouted, renamed or materially reconfigured while a study is collecting data, under review or in production.

The resulting problem is partly empirical and partly grammatical. ``GPT-4o, queried in September 2024, achieved 68\%'' is a dated observation. ``Large language models achieve 68\%'' turns the same result into a present-tense claim about a changing population. The experiment may remain valid while the sentence outgrows it.

Calendar age is not itself a measure of quality or validity. A later model may perform similarly, and an older result may remain the best evidence about the system, people or institution actually studied. Conversely, a recent posting can evaluate a model family that is already several generations behind. The important question is how far a conclusion can travel beyond the system and date from which it was derived.

This paper uses \textbf{model age} for the elapsed time between release of the newest named model generation or immutable snapshot and the record's appearance. It uses \textbf{claim currency} for the extent to which a conclusion remains appropriately supported after changes in models, access routes, tools or surrounding systems. The distinction matters because claims do not age uniformly. A current capability ranking may move after one successor release; a finding about student trust, clinical workflow or institutional adoption may remain informative for much longer.

This paper is deliberately dual-purpose. Its first aim is substantive: to identify a recurring weakness in generative-AI evidence and develop a practical response. Its second is reflexive: to provide an inspectable example of a single researcher using a frontier model across much of the research-creation process, rather than using AI only to polish prose. Here, \emph{frontier} denotes a recently released, high-capability general-purpose system available at the time of the work; it does not mean that the system was established as best on every task. The production case is evidence about one workflow, not a general productivity result.

The audit addresses three research questions:

\par\noindent\hangindent=1.7em\hangafter=1\textbf{1.}\ How old was the newest named model when each included record appeared?\par
\par\noindent\hangindent=1.7em\hangafter=1\textbf{2.}\ How often did records provide a precise model identifier, include a family superseded by appearance, or refresh evidence with a successor model?\par
\par\noindent\hangindent=1.7em\hangafter=1\textbf{3.}\ What reporting practices could help readers interpret and maintain claims about changing generative-AI systems?\par
The empirical contribution is a fully inspectable, record-level audit of 40 publications and preprints across one year. The conceptual contribution is the move from paper-level age to claim-level currency. The practical contribution is a lifecycle for reporting: identify the tested system, date the claim, record changes at acceptance, bridge sensitive claims when justified, and preserve updates as versioned additions rather than silent replacements. The reflexive contribution is a self-applying production case: the paper reports Model Facts, a completed model-currency statement and the controls used when AI assisted discovery, reconciliation, calculation, drafting and criticism.

\section{Methods}
\subsection{Design and scope}
The study is a descriptive, maximum-variation purposive audit. It was developed iteratively rather than from a preregistered search protocol. The one-year inclusion window ran from 18 July 2025 to 17 July 2026. Records were selected to cover a wide range of tasks and publication routes, including clinical and professional decision-making, education and assessment, social behaviour, coding, reasoning, model judging, routing, multimodal tasks and agentic systems.

Forty records was a practical stopping point. It allowed recurring patterns to be examined while keeping the full coding visible in Appendix A and in the accompanying comma-separated data file. The corpus is not a random or exhaustive sample and cannot support field-wide prevalence estimates.

\subsection{Search and eligibility}
Candidate records were generated through AI-assisted searching and then opened, checked, selected and coded by the author. Sources included journal and publisher sites, bibliographic search results, arXiv and provider release pages. Searching was redirected as gaps in discipline, task, model family and publication route became visible, and it continued to the data cutoff.

A record was eligible when it:

\begin{itemize}[leftmargin=1.6em,itemsep=0.2em]
\item reported original empirical testing of at least one named generative-AI model;
\item had a verifiable first-online, release or audited-version date within the inclusion window; and
\item provided enough information to estimate the age of the newest tested generation or immutable snapshot.
\end{itemize}
Reviews, commentaries, duplicate records, non-generative systems and records without adequate model or timing information were excluded. Model age did not determine inclusion.

There was no fixed search string, exhaustive screening log, independent second screener or formal risk-of-bias assessment. Selection and coding rules were refined during the inquiry. These limitations are material: the corpus supports descriptive comparison and framework development, not claims about the prevalence or quality of generative-AI research as a whole.

\subsection{Unit of analysis and coding rules}
The unit of analysis was the record as it appeared within the audit window: first-online publication for journal articles, the audited version for preprints, and release date for the laboratory report. Table 1 summarises the coded fields and operational rules.

\begin{table}[htbp]
\centering
\caption{Coding fields and operational rules.}
\label{tab:coding}
\footnotesize
\begin{tabularx}{\textwidth}{>{\RaggedRight\arraybackslash}p{0.19\textwidth} X X}
\toprule
\textbf{Field} & \textbf{Operational rule} & \textbf{Qualification} \\
\midrule
Appearance date & First-online date for journal articles; audited version date for preprints; release date for the laboratory report. & May differ from issue year or later preprint versions. \\
Newest named system & Most recent named generation or immutable snapshot tested in the record. & May not be the principal intervention or the model used for most comparisons. \\
Model age & Calendar days from public release of that generation or snapshot to record appearance. & For floating labels, family release may not identify the backend served. \\
Precise identifier & Dated API snapshot or another designation with comparable build-level specificity. & A family or consumer-product name alone was not treated as precise. \\
Supersession & A public same-family successor was available by the appearance date. & Does not imply retirement, invalidity or improved performance on the studied task. \\
Refresh & An explicit rerun of some or all evaluation with a successor after original testing. & A later model added without update framing was coded as a late sensitivity test. \\
Route and task & Retained as descriptive attributes. & Not treated as causal predictors. \\
\bottomrule
\end{tabularx}
\end{table}

Model age was calculated as the number of calendar days between release of the newest named generation or immutable snapshot and the record's appearance date. For floating product labels such as ChatGPT, Copilot or GPT-4o, the public family-release date was used. This may not identify the backend served during execution. The measure is also conservative in one respect: it uses the newest system in a panel even when most analyses or the principal intervention relied on older models.

A \textbf{precise identifier} was a dated API snapshot or another model designation that fixed the tested build with comparable specificity. \textbf{Supersession} meant that a public same-family successor was available by the appearance date. It did not mean that the older model had been withdrawn or that the result was invalid. A \textbf{clear refresh} required an explicit rerun of some or all of the evaluation with a successor model after the original testing. A newer model added without being described as an update to the original evidence was recorded separately as a late sensitivity test.

\subsection{Analysis and verification}
The analysis is descriptive. It reports counts, medians, middle-50\% ranges and observed ranges; no inferential statistics were used. Publication route and task were retained as descriptive attributes rather than treated as causal predictors. Summary statistics were recalculated from the Appendix A coding during final verification. The age values reproduce the reported median of 281 days, middle-50\% range of 75--478 days, and route-specific medians.

Release dates were checked against public provider records available by 17 July 2026. Corpus references and paper-level coding are published in full so that readers can inspect the decisions and revise them if better information becomes available. The claim-sensitivity taxonomy and reporting framework developed from the audit are proposals; they have not yet been validated as an expiry schedule or tested against downstream decision quality.

\subsection{AI-assisted production as a reflexive case}
The paper was developed over two days through repeated cycles of question refinement, candidate discovery, source checking, date reconciliation, calculation, drafting, criticism and document production involving the author and GPT-5.6 Sol Pro in ChatGPT. The system was used as a research instrument across the workflow, not only as a language editor. It proposed records and lines of inquiry, helped compare dates and sources, ran or checked calculations, generated alternative interpretations, drafted passages and challenged the developing argument.

The author set the question and inclusion boundaries, opened and evaluated sources, made every selection and coding decision, checked numerical summaries, rewrote the manuscript and accepts responsibility. AI-generated criticism was not independent validation. No participant-level or confidential data were used.

The production case was documented retrospectively and is offered as a proof-of-practice at the frontier of AI-assisted research creation. It has no unaided comparator, prospective time log, immutable model snapshot or independent assessment of the workflow. Exact session timestamps, backend routing, default settings, token use and cost were not prospectively recorded. It can show what was assembled under this human-AI arrangement; it cannot establish how much faster or better the work was.

\section{Results}
\subsection{Model age at appearance}
The corpus comprised 25 journal articles, 14 preprints and one laboratory report. At appearance, the newest named model was a median 281 days old. The middle half of records fell between 75 and 478 days, and the full range was 11 to 939 days (Table 2).

\begin{table}[htbp]
\centering
\caption{Headline findings from the 40-record corpus. Counts describe this purposive collection and are not prevalence estimates for generative-AI research.}
\label{tab:headline}
\footnotesize
\begin{tabularx}{\textwidth}{>{\RaggedRight\arraybackslash}p{0.22\textwidth} X X}
\toprule
\textbf{Measure} & \textbf{Finding in the corpus} & \textbf{Interpretation} \\
\midrule
Overall model age & Median 281 days; middle 50\%: 75--478; range: 11--939. & The newest named model was typically about nine months old when a record appeared. \\
Supersession at appearance & 35 of 40 records included at least one tested family with a public same-family successor. & Most records were historical in at least one model family by appearance. \\
Precise model identity & 7 of 40 supplied a dated API snapshot or comparably precise identifier. & Labels such as ChatGPT, GPT-4o or Copilot often left the tested backend uncertain. \\
Evidence refresh & 3 clear refreshes; 1 separate late sensitivity test. & Few records connected an original evaluation to a successor model. \\
Provider concentration & 40 of 40 included an OpenAI system. & This is a feature of the corpus, not an estimate of field-wide market share. \\
Publication route & 25 journal articles: median 395 days; 14 preprints: median 56 days; 1 laboratory report: 49 days. & Preprints were closer to releases at posting, but route did not guarantee currency. \\
\bottomrule
\end{tabularx}
\end{table}

\begin{figure}[htbp]
\centering
\includegraphics[width=0.95\textwidth]{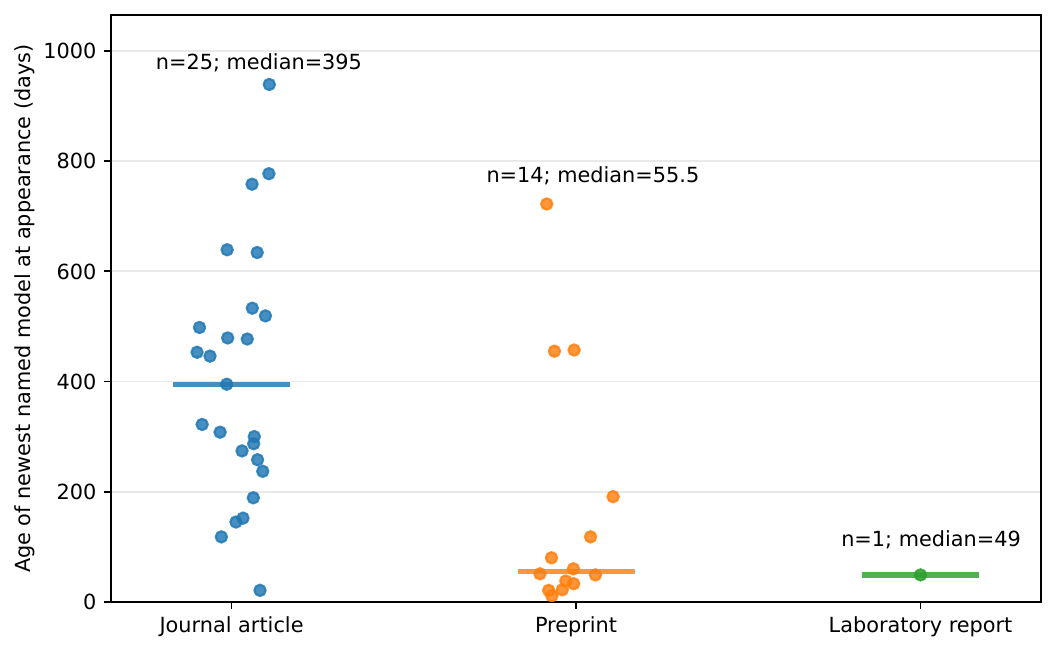}
\caption{Age of the newest named model at record appearance, by publication route. Each point is one record; horizontal marks show medians. The single laboratory report is displayed separately. The figure describes this purposive corpus and does not estimate route-level population differences.}
\label{fig:modelage}
\end{figure}

Publication route described a substantial difference within this corpus but did not determine currency. Journal articles had a median model age of 395 days; preprints had a median of 56 days; the laboratory report was 49 days. Several journal articles added relatively recent systems during revision. By contrast, three preprints posted on 15 July 2026 used newest model families or snapshots estimated to be 455--722 days old. A recent posting date is therefore not a proxy for recent model evidence.

\subsection{Identification, supersession and refresh}
Seven of the 40 records supplied a dated API snapshot or comparably precise identifier. The remaining records used family names, consumer-product labels or model descriptions that left some uncertainty about the exact system served. This was especially important for mutable interfaces and routed products, where the same label may not map to a stable backend.

Thirty-five records included at least one tested family for which a public same-family successor was available by the appearance date. Supersession was common even among newly posted preprints. It does not negate the original result; it changes the wording and evidential burden when authors generalise from a historical panel to the present field.

Only three records clearly refreshed part or all of their model evidence with a successor system. One further record added a later validation that functioned as a sensitivity test but was not framed as a refresh. Most records therefore preserved the original experiment without showing whether its central conclusion survived a material model change.

Every included record tested an OpenAI system. This concentration may reflect both research practice and the audit's search pathway. It must not be interpreted as a market-share estimate. A much larger clinical review of 4,609 studies found ChatGPT and related OpenAI systems in 65.7\% of the evaluated models, which independently shows substantial provider concentration in at least one domain \cite{ref14}.

\section{From model age to claim currency}
\subsection{Three forms of distance}
The audit measures \textbf{calendar distance}: elapsed days between model release and record appearance. Two further distances are relevant but were not coded.

\textbf{Generational distance} concerns the number and scale of successor releases since testing. Two models that are each 200 days old may sit very different numbers of releases behind their respective families. \textbf{System distance} concerns changes around the model: reasoning effort, search, retrieval, tools, memory, context handling, routing, scaffolding, judge models and multi-agent orchestration. The same model label can therefore denote materially different interventions.

The public frontier at the cutoff illustrates the problem. GPT-5.6 had been released on 9 July 2026 \cite{ref1}. Kimi K3 and Grok 4.5 were announced on 16 July \cite{ref6,ref7}. Access to Claude Fable 5 had been restored on 1 July after its June launch, Claude Sonnet 5 was released on 30 June, Gemini 3.5 Flash on 19 May, and DeepSeek V4 Preview on 24 April \cite{ref2,ref3,ref4,ref5,ref8}. This chronology is included only to date the audit context; it is not a capability ranking. Later releases will make it historical without altering what was true at the cutoff.

Names may also conceal routing. Anthropic reported that safeguard-triggering Fable 5 requests could be routed to Opus 4.8 \cite{ref2,ref3}. Kimi described K3 as a 2.8-trillion-parameter sparse mixture-of-experts system activating 16 of 896 experts while full weights and a technical report were still forthcoming \cite{ref6}. These examples show why a parameter count or product name cannot, by itself, identify the intervention. Access route, tools, reasoning budget, system instructions and fallbacks can be equally consequential.

\subsection{Why the age of evidence can change a conclusion}
Age is not an effect size, and no universal penalty can be inferred from days alone. A successor may barely alter performance on a saturated task yet cross an operational threshold in coding, clinical reasoning or agentic work. Provider evaluations offer dated illustrations of that possibility, although they combine changes in models, prompts, tools and evaluation harnesses and must not be treated as a common-harness estimate (Table 3).

\begin{table}[htbp]
\centering
\caption{Dated provider comparisons illustrating why claim currency can matter. The evaluations used different prompts, tools and harnesses and do not estimate a general effect of model age.}
\label{tab:comparisons}
\scriptsize
\begin{tabularx}{\textwidth}{>{\RaggedRight\arraybackslash}p{0.17\textwidth} X >{\RaggedRight\arraybackslash}p{0.17\textwidth} X}
\toprule
\textbf{Illustration} & \textbf{Reported comparison} & \textbf{Observed difference} & \textbf{Relevance to claim currency} \\
\midrule
SWE-bench Verified & GPT-4o 33.2\%; GPT-5 74.9\% \cite{ref9,ref10,ref11}. & +41.7 percentage points. & A successor can cross a practical coding threshold, although the surrounding evaluation also changed. \\
Factual-error rate & GPT-5 was reported to make about 45\% fewer factual errors than GPT-4o on production-like web-search queries \cite{ref12}. & About 45\% lower reported error rate. & Successor systems can alter error profiles, not only aggregate scores. \\
GPT-5.5 to GPT-5.6 & Reported gains included +5.8 on Agents' Last Exam, +11.0 on HealthBench Professional, +15.1 on OSWorld 2.0 and +6.0 on BrowseComp \cite{ref1}. & +5.8 to +15.1 points. & Even a same-family update can change rankings or workflow feasibility. \\
\bottomrule
\end{tabularx}
\end{table}

The relevant question is therefore not ``How much should an old result be discounted?'' It is ``Which part of the conclusion depends on proximity to the current system?'' Absolute accuracy, rankings, feasibility thresholds, refusal behaviour, latency, cost and claims that a system can or cannot perform a task are often highly sensitive to successor releases or product changes. Findings about people, institutions and workflow may remain informative for longer, although the tested model remains part of the intervention.

Table 4 presents a claim-level reporting taxonomy. It is a proposal for deciding what deserves review or bridging, not a validated expiry schedule.

\begin{table}[htbp]
\centering
\caption{Proposed claim-sensitivity taxonomy. This is a reporting aid, not a validated expiry schedule.}
\label{tab:claims}
\scriptsize
\begin{tabularx}{\textwidth}{>{\RaggedRight\arraybackslash}p{0.18\textwidth} X X X}
\toprule
\textbf{Claim type} & \textbf{Examples} & \textbf{Likely sensitivity} & \textbf{Minimum reporting or update action} \\
\midrule
Current capability or ranking & Accuracy, pass rates, leaderboard position, \texttt{}AI can/cannot claims. & High after a major successor or material harness change. & Model Facts, dated grammar, acceptance frontier note and a justified bridge when the conclusion may move. \\
Deployed-system behaviour & Refusals, routing, tool use, latency and cost. & High when product, access route, policy or harness changes. & Report the full configuration, observable routing and refusals, plus a material-change trigger. \\
Error patterns and mechanisms & Failure categories, prompt effects, calibration and judge behaviour. & Variable; the pattern may persist even when the score changes. & Retest representative failures or threshold cases and state which mechanism is expected to generalise. \\
Human and institutional effects & Trust, learning, adoption, workflow and assessment. & Often slower, but not independent of the model intervention. & Preserve model details and assess whether the proposed mechanism depends on capability or product design. \\
\bottomrule
\end{tabularx}
\end{table}

A dated result remains evidence about the system that was actually tested. What can shrink is the scope of inference. This is why the unit of refresh should be the claim rather than the paper.

\section{A model-currency reporting framework}
A full rerun after every release would be costly, disruptive and often unnecessary. Update effort should instead follow the sensitivity, stakes and intended reach of the central claim. The proposed framework has six parts.

\par\noindent\hangindent=1.7em\hangafter=1\textbf{1.}\ \textbf{State the claim's currency requirement.} Identify whether the paper makes a point-in-time observation, a current capability or ranking claim, a deployed-system behaviour claim, or a human or institutional finding expected to outlast one model generation. This classification should determine the update burden.\par
\par\noindent\hangindent=1.7em\hangafter=1\textbf{2.}\ \textbf{Publish Model Facts.} Record the exact model or snapshot, provider, access route, region, execution dates, system prompt, sampling settings, reasoning effort, tools, retrieval, context handling, repetition count, judge model and cost where relevant. A consumer label such as ``ChatGPT'' is not a reproducible model description.\par
\par\noindent\hangindent=1.7em\hangafter=1\textbf{3.}\ \textbf{Add a frontier note at acceptance.} List material same-family releases and system changes since the data freeze. State which conclusions they might affect and why. A dated note is informative even when no rerun is possible.\par
\par\noindent\hangindent=1.7em\hangafter=1\textbf{4.}\ \textbf{Predefine a sentinel bridge for sensitive claims.} Where a central claim is both likely to move and consequential, select a small, justified subset of cases for successor testing. The subset should cover the parts of the task most capable of changing the conclusion, such as threshold cases, difficult items, important failure modes or safety-critical examples. Its size and composition should be justified by the task and stakes rather than a universal percentage.\par
\par\noindent\hangindent=1.7em\hangafter=1\textbf{5.}\ \textbf{Preserve the original analysis and version the update.} The planned or registered experiment should remain intact. Report bridge results in a dated supplement or new version, including outcomes that weaken, reverse or leave the original conclusion unchanged. An update should extend the evidence trail, not rewrite it.\par
\par\noindent\hangindent=1.7em\hangafter=1\textbf{6.}\ \textbf{Set a refresh trigger and use dated grammar.} Triggers may include a major same-family successor, retirement of the tested model, a material change in tools or routing, or a prespecified review date. Write ``snapshot X, run on date Y, achieved Z under configuration C'' rather than ``LLMs achieve Z''. Report refusals, fallbacks and routing when observable.\par
These practices extend reporting guidance such as TRIPOD-LLM by adding a lifecycle for claims whose object of study can change after data collection \cite{ref13}. They make the time boundary of an inference explicit while preserving the original result.

\begin{quote}
\small
\textbf{Suggested model-currency statement.} Experiments ran between [dates] using [exact model identifiers, access route and system configuration]. The reported results apply to those systems under the stated conditions. The central claim is a [point-in-time observation / current capability or ranking / deployed-system behaviour / longer-lived human or institutional finding]. Since the data freeze, [successor models or material system changes] have appeared. We [did not rerun / reran] a prespecified sentinel subset of [cases and rationale] with [new identifiers] on [date]. The conclusion was [stable / weaker / changed]. Routing, refusals and fallback models were recorded where observable. The next review is due [date or trigger], or no further refresh is planned.
\end{quote}

\subsection{Applying the framework to this paper}
A reporting framework is more credible when the paper proposing it accepts the same burden. Table~\ref{tab:selffacts} records Model Facts for the AI-assisted production process. The completed statement that follows applies the generic template above to this manuscript.

\begin{table}[htbp]
\centering
\caption{Model Facts for the AI-assisted production process.}
\label{tab:selffacts}
\footnotesize
\begin{tabularx}{\textwidth}{>{\RaggedRight\arraybackslash}p{0.24\textwidth} X}
\toprule
\textbf{Field} & \textbf{Reported information} \\
\midrule
Research purpose and status & Substantive 40-record audit and reporting framework, accompanied by a reflexive proof-of-practice. The production case is descriptive, not a controlled productivity or quality comparison. \\
System label & GPT-5.6 Sol Pro, as displayed to the author in ChatGPT. \\
Frontier status & GPT-5.6 was released on 9 July 2026, eight days before the corpus cutoff \cite{ref1}. ``Frontier'' denotes a newly released, high-capability general-purpose model, not best-on-every-task status. \\
Provider and access & OpenAI through the ChatGPT product interface. An immutable backend identifier and routing path were not exposed to the author. \\
Production period & Two-day production period before manuscript freeze. Exact session timestamps were not prospectively logged. \\
AI roles & Candidate discovery, search direction, date and source reconciliation, calculations, alternative interpretations, drafting, criticism and document production. \\
Human controls & The author set the question and rules, opened and evaluated sources, made coding and interpretive decisions, checked numerical summaries, rewrote the manuscript and accepts responsibility. \\
Data and privacy & Public research records and provider release pages. No participant-level or confidential data were used. \\
Unobserved or unlogged & Exact backend, routing, default settings, system instructions, token use and cost. \\
Reproducibility and artefacts & The record-level corpus and coding are published in Appendix A and \texttt{anc/corpus\_coding.csv}. The exact interaction cannot be reconstructed from the product label alone. \\
\bottomrule
\end{tabularx}
\end{table}

\begin{quote}
\small
\textbf{Completed model-currency statement for this paper.} Corpus identification, checking and coding were frozen on 17 July 2026. The AI-assisted production process used GPT-5.6 Sol Pro through ChatGPT over two days; exact session timestamps, an immutable backend identifier, routing, default settings, token use and cost were not prospectively recorded. The numerical findings are point-in-time descriptions of the 40-record corpus. The claim-currency framework and production case are methodological propositions rather than claims that this model or workflow is currently best. By 27 July 2026, Google DeepMind had published Gemini 3.6 Flash and introduced Gemini 3.5 Flash Cyber on 21 July \cite{ref25,ref26}. These developments post-date the corpus freeze and do not alter any calculation or the central framework. No sentinel rerun was conducted because the audit findings depend on source records and coding, not GPT-5.6's benchmark performance. The production account remains a descriptive case without an unaided comparator. Review is triggered by an expanded corpus, material evidence that changes the framework, or 17 July 2027, whichever comes first.
\end{quote}

\section{Discussion}
\subsection{What the audit shows---and what it does not}
The audit documents a recurring mismatch between the pace of model change and the pace of publication. In this corpus, the newest named model was typically about nine months old when a record appeared, and at least one tested family had already been superseded in most records. Exact system identification and explicit refreshes were uncommon.

Those observations do not establish that journal evidence is generally obsolete, that preprints are generally current, or that performance decays at a predictable rate. The sample is purposive, and model age alone says nothing about task difficulty, benchmark quality, effect size or the durability of a mechanism. A 500-day-old study may remain authoritative about a historical intervention. A 20-day-old benchmark may already be misleading if it uses a routed product, an unreported harness or a claim stated at the level of ``AI''.

Publication delay becomes a scientific problem when the wording of a conclusion implies present capability while its evidence is tied to a materially older system. The remedy is not to discard slower scholarship. It is to distinguish historical measurement from current inference and to update only the claims that genuinely depend on the frontier.

\subsection{A layered publication model}
Preprints were much closer to model releases at posting in this corpus, and their versioning makes them a useful home for dated capability evidence and bridge tests. That advantage is conditional. Rapid posting does not guarantee sound benchmark design, statistical analysis, prompts, judge models or interpretation, and a new preprint may test an old model.

A layered sequence is more useful than a competition between routes. Authors can release a dated preprint, code and machine-readable evidence for checking and later updates, while journal review develops claims intended to endure. The preprint can carry a current evidence layer; the journal can provide deeper appraisal of design, interpretation, ethics and consequences. The audit does not measure whether the present reviewer pool has the specialist capacity required for this arrangement. Separate work documents wider scale pressures in machine-learning peer review \cite{ref21}.

\subsection{Relationship to prior work}
The closest identified study is Gringras and Salahshoor's \emph{Frontier Lag}, version 2 \cite{ref15}. Its preregistered OpenAlex audit covered more than 18,000 admissible records through 1 April 2026 and used benchmark-composite indices to estimate capability distance from the frontier. It reported a substantial median capability gap, sparse reporting of reasoning mode, and frequent claims made at the level of ``AI'' rather than the tested model. That study establishes the scale of frontier lag. The present paper contributes a smaller, inspectable audit in calendar days and asks how the currency of particular claims should be reported and maintained.

Related work places the problem within a broader reproducibility deficit. Balloccu et al. examined contamination and evaluation malpractice in closed-source LLM studies \cite{ref16}. Ko et al. assessed 159 healthcare studies against MI-CLEAR-LLM \cite{ref17}. Evertz et al. found at least one methodological pitfall in each of 72 security and software-engineering papers \cite{ref18}. ReproEvalCard reported missing randomness controls in 75\% of 55 LLM-pipeline papers and missing intermediate traces in 61\% \cite{ref19}. A 640-paper software-engineering audit found persistent gaps in artefacts, versioning and model documentation \cite{ref20}. Model currency is not a substitute for those controls. It is an additional lifecycle problem: even a well-designed and reproducible evaluation can be misread if a dated intervention is expressed as a timeless property of ``AI''.

\subsection{Implications for research and publication}
The framework suggests several practical changes. Evidence syntheses can code both model age and claim type rather than treating publication year as a sufficient indicator of currency. Editors can request a short frontier note at acceptance for papers whose central conclusions concern current capability or deployed-system behaviour. Authors can retain preregistered analyses while adding a versioned bridge, avoiding the choice between methodological fidelity and technological relevance. Repositories can publish machine-readable model facts and dated supplements alongside the article.

The framework also creates testable questions. Future work should determine whether claim-sensitivity classifications can be applied consistently across reviewers; whether small sentinel bridges predict the direction and magnitude of full reruns; which releases or system changes should count as material triggers; and how calendar, generational and system distance jointly relate to changes in conclusions. Larger, protocol-driven samples are needed to estimate prevalence by discipline, provider and publication route.

\subsection{Frontier-model-assisted research creation}
AI-assisted research creation already spans a wide range. Bubeck et al. report human-guided cases in which GPT-5 contributed to literature review, hypothesis development, calculations and proofs \cite{ref22}. Swanson et al. describe a human-supervised virtual laboratory of AI agents whose computational designs were tested experimentally \cite{ref23}. Lu et al. present an agentic system intended to automate much of the machine-learning research cycle \cite{ref24}. The present paper sits at the human-led end of that spectrum: one researcher retained epistemic and editorial control while a frontier model was used repeatedly across discovery, reconciliation, calculation, interpretation, drafting and critique.

The point of the case is not simply that AI helped write a paper in two days. Text generation is the least distinctive part. The more consequential possibility is that a single researcher can traverse a larger search and analytic space, test more interpretations and assemble inspectable evidence artefacts more quickly than would otherwise be practical. In this audit, the process helped develop the distinction between model age and claim currency and turn it into a record-level corpus and reporting framework. Because there is no comparator, this is an example to investigate rather than a measured productivity gain.

The production process also mirrors the paper's object of study. The AI system was available through a mutable product whose exact backend and routing were not independently observable. A future reader cannot reproduce the interaction from the product label alone. Model Facts and the completed currency statement make those limits explicit, but do not eliminate them.

Fast iteration can transmit errors quickly. An incorrect release date, weak inference or invented source can move from search to calculation to prose, and AI criticism of AI-assisted work is not independent validation. Visible coding, source trails, recalculated summaries and accountable human review are necessary controls. They do not remove the need for external peer review.

\subsection{Limitations}
The corpus is purposive, not exhaustive or statistically representative. Manual selection may favour visible publishers, English-language work and studies with enough detail to audit. Every record included an OpenAI system, which may reflect both field practice and selection. The counts apply only to these 40 records.

Model age is approximate and incomplete. Family-release dates cannot reveal mutable backends, and using the newest model in a panel may make an older principal experiment appear more current. The audit measures calendar distance but not generational or system distance. It does not assess study quality, benchmark validity, statistical power or the substantive importance of model changes.

The provider comparisons are illustrative and use different prompts, tools and harnesses. They cannot estimate an effect of age. Coding was conducted by one author without independent screening or inter-rater reliability. The claim taxonomy and reporting framework are proposals derived from the audit, not validated standards. The AI-production account concerns one researcher, one model-assisted manuscript and two days of work, with no comparator, prospective log or independent workflow assessment. It cannot establish productivity, originality, quality, causal contribution or generalisability. The author-model division of labour is also reconstructed from the completed process rather than measured prospectively.

The data cutoff is 17 July 2026. Later releases do not change the paper-level calculations, but they will make the frozen frontier context historical. The appendix and accompanying data file preserve what was measured and when so that later readers can inspect or update it.

\section{Conclusion}
In this purposive corpus of 40 empirical records, the newest named model was a median 281 days old when the record appeared. Thirty-five records included a superseded family, seven supplied a dated snapshot or comparably precise identifier, three clearly refreshed their model evidence, and one added a late sensitivity test. Journal articles were older than preprints at appearance in this sample, but neither route guaranteed currency.

The result is not a universal half-life for generative-AI evidence. Different claims carry different update burdens. Rankings, capability thresholds and statements about what a current system can or cannot do are likely to move quickly. Findings about people, workflow and institutions may travel further, although the model remains part of the intervention.

Researchers can make that boundary visible. Name the system and configuration, state when it was tested, classify the claim's currency requirement, record material changes at acceptance, bridge sensitive claims when justified, and publish updates as dated versions. Scholarship cannot eliminate the mismatch between fast-changing systems and slower review. It can stop that mismatch from being hidden in timeless language.

The paper also advances a second, narrower proposition. Frontier AI can participate in research creation as more than a writing aid: under human direction it can widen discovery, reconciliation, analysis and criticism. This single case does not prove a productivity gain, autonomous research capability or superior quality. It provides a worked, self-disclosing example whose value depends on visible sources, reproducible calculations, clear responsibility and external scrutiny.

\section*{Declarations}
\textbf{Author contributions.} Carlo Iacono: conceptualisation, methodology, investigation, data curation, formal analysis, visualisation, writing---original draft, writing---review and editing, and project administration. The author made all final decisions and accepts responsibility for the work.

\textbf{Ethics.} This audit analysed published and publicly available research records. It did not recruit human participants or collect individual-level participant data.

\textbf{Data availability.} The 40 included records and the paper-level coding used for the descriptive summaries are reported in Appendix A and in the accompanying \texttt{corpus\_coding.csv} file. Because the collection was purposive and the search developed during the inquiry, it is not an exhaustive literature dataset and no exhaustive screening log exists.

\textbf{Competing interests.} The author declares no competing interests.

\textbf{Generative-AI use.} GPT-5.6 Sol Pro was accessed through ChatGPT during the two-day production period as a research instrument across candidate discovery, evidence gathering, date and source reconciliation, calculations, alternative interpretation, drafting, revision, criticism and document production. Exact session timestamps and an immutable backend snapshot were not prospectively recorded, so the underlying build, routing and default settings were not independently observable. The author repeatedly reviewed, challenged, redirected and edited the outputs, made every inclusion and interpretive decision, and decided when to stop. AI critique formed part of production and was not independent validation. Table~\ref{tab:selffacts} and the completed model-currency statement report the workflow's known facts and limits. The author accepts responsibility for the sources, analysis, claims and final text.

The references below support release dates, illustrative comparisons, the reporting framework, the submission-stage frontier note and related work on AI-assisted research. The audited empirical records are listed separately as C1--C40 in Appendix A.

\appendix
\section{Corpus references and paper-level coding}
\subsection{Corpus references (C1--C40)}
The records are listed in the same chronological order as the coding table. Audit dates are first-online, release or version dates and may differ from the formal citation year.

\noindent\textbf{C1.} Bedi S, Jiang Y, Chung P, Koyejo S, Shah N. Fidelity of Medical Reasoning in Large Language Models. JAMA Network Open. 2025;8(8):e2526021. doi:10.1001/jamanetworkopen.2025.26021.\par
\noindent\textbf{C2.} Zhang Y, Li H, Chen J, et al. Beyond GPT-5: Making LLMs Cheaper and Better via Performance-Efficiency Optimized Routing. In: DAI '25: Proceedings of the 2025 7th International Conference on Distributed Artificial Intelligence. 2025:122-129. doi:10.1145/3772429.3772445; arXiv:2508.12631v2.\par
\noindent\textbf{C3.} Shyr C, Cassini TA, Tinker RJ, et al. Large Language Models for Rare Disease Diagnosis at the Undiagnosed Diseases Network. JAMA Network Open. 2025;8(8):e2528538. doi:10.1001/jamanetworkopen.2025.28538.\par
\noindent\textbf{C4.} Kliem PSC, Fisch U, Baumann SM, et al. The impact of prompting on ChatGPT's adherence to status epilepticus treatment guidelines. Scientific Reports. 2025;15:31712. doi:10.1038/s41598-025-16902-9.\par
\noindent\textbf{C5.} Xiao F, Wang XX. Evaluating the ability of large language models to predict human social decisions. Scientific Reports. 2025;15:32290. doi:10.1038/s41598-025-17188-7.\par
\noindent\textbf{C6.} Liu S, Huang SS, McCoy AB, et al. Optimizing Order Sets With a Large Language Model-Powered Multiagent System. JAMA Network Open. 2025;8(9):e2533277. doi:10.1001/jamanetworkopen.2025.33277.\par
\noindent\textbf{C7.} OpenAI. Measuring the performance of our models on real-world tasks. OpenAI Publication. 2025. Related paper: Patwardhan T, Dias R, Proehl E, et al. GDPval: Evaluating AI Model Performance on Real-World Economically Valuable Tasks. arXiv:2510.04374v1.\par
\noindent\textbf{C8.} Pinheiro LCD, Chen Z, Piazza BC, et al. Large Language Models Achieve Gold Medal Performance at the International Olympiad on Astronomy and Astrophysics (IOAA). arXiv. 2025. arXiv:2510.05016v2.\par
\noindent\textbf{C9.} Zubaer AA, Granitzer M, Geschwind S, et al. GPT-4 shows comparable performance to human examiners in ranking open-text answers. Scientific Reports. 2025;15:35045. doi:10.1038/s41598-025-21572-8.\par
\noindent\textbf{C10.} Gholami S, Mummert DB, Wilson B, et al. Leveraging Large Language Models to Generate Multiple-Choice Questions for Ophthalmology Education. JAMA Ophthalmology. 2025;143(11):955-961. doi:10.1001/jamaophthalmol.2025.3622.\par
\noindent\textbf{C11.} Arora V, Thabane A, Parpia S, et al. Generative artificial intelligence models outperform students on divergent and convergent thinking assessments. Scientific Reports. 2025;15:36987. doi:10.1038/s41598-025-21398-4.\par
\noindent\textbf{C12.} Idan D, Ben-Shitrit I, Volevich M, et al. Evaluating the performance of large language models versus human researchers on real world complex medical queries. Scientific Reports. 2025;15:37824. doi:10.1038/s41598-025-21689-w.\par
\noindent\textbf{C13.} Suresh VK. Two-Faced Social Agents: Context Collapse in Role-Conditioned Large Language Models. arXiv. 2025. arXiv:2511.15573v1.\par
\noindent\textbf{C14.} Skjervold K, Saevig HN, Raeder H, et al. DiaGuide-LLM: Using large language models for patient-specific education and health guidance in diabetes. Frontiers in Artificial Intelligence. 2025;8:1652556. doi:10.3389/frai.2025.1652556.\par
\noindent\textbf{C15.} Sheikhalishahi S, Haddadi A, Sadeghipour S, et al. Comparative performance of ChatGPT-4o, ChatGPT-5, and Gemini 2.5 Flash on Persian internal medicine subspecialty board exams. Scientific Reports. 2026;16:1371. doi:10.1038/s41598-025-31251-3.\par
\noindent\textbf{C16.} Patel J, Chen Y, He K, et al. Reasoning Models Ace the CFA Exams. arXiv. 2025. arXiv:2512.08270v1.\par
\noindent\textbf{C17.} Feng Y, Wang S, Cheng Z, Wan Y, Chen D. Are We on the Right Way to Assessing LLM-as-a-Judge? arXiv. 2025. arXiv:2512.16041v1.\par
\noindent\textbf{C18.} Bai H, Lui WC, Khiatani PV. Promoting student engagement with GPTutor: An intelligent tutoring system powered by generative AI. International Journal of Educational Technology in Higher Education. 2025;22:77. doi:10.1186/s41239-025-00571-9.\par
\noindent\textbf{C19.} Patel A, Contractor H, Heninger H, et al. Performance of successive generative pretrained transformers (GPT) models in medical cases and board style questions. Scientific Reports. 2026;16:4782. doi:10.1038/s41598-025-34939-8.\par
\noindent\textbf{C20.} Yu Z, Zhou C, Lin Y, et al. ChipBench: A Next-Step Benchmark for Evaluating LLM Performance in AI-Aided Chip Design. arXiv. 2026. arXiv:2601.21448v2.\par
\noindent\textbf{C21.} Sajja R, Sermet Y, Fodale B, et al. Evaluating AI-powered learning assistants in engineering higher education with implications for student engagement, ethics, and policy. Scientific Reports. 2026;16:7565. doi:10.1038/s41598-026-39237-5.\par
\noindent\textbf{C22.} Cheung KKC, Zhang W. Representations of the Nature of Science in Generative AI (GPT-4o). Science \& Education. 2026. doi:10.1007/s11191-025-00718-0.\par
\noindent\textbf{C23.} Oh J, Whang SE, Evans J, Wang J. Classroom AI: large language models as grade-specific teachers. npj Artificial Intelligence. 2026;2:28. doi:10.1038/s44387-026-00081-7.\par
\noindent\textbf{C24.} Tu M, Ni S, Bi K. How Long Reasoning Chains Influence LLMs' Judgment of Answer Factuality. arXiv. 2026. arXiv:2604.06756v1.\par
\noindent\textbf{C25.} Rao AS, Esmail KP, Lee RS, et al. Large Language Model Performance and Clinical Reasoning Tasks. JAMA Network Open. 2026;9(4):e264003. doi:10.1001/jamanetworkopen.2026.4003.\par
\noindent\textbf{C26.} Saez Y, Garcia LM, Mochon A, et al. Evaluating large language models for AI-assisted grading: a framework and case study in higher education. Scientific Reports. 2026;16:18035. doi:10.1038/s41598-026-48656-3.\par
\noindent\textbf{C27.} Zhang Y, Chen H, Xu R, et al. Benchmarking clinical knowledge and multi-modal reasoning of large language models in liver cirrhosis. Scientific Reports. 2026;16:18722. doi:10.1038/s41598-026-44685-0.\par
\noindent\textbf{C28.} Sheikhalishahi S, Rafiei F, Hosseini SM, et al. Benchmarking large language models on Persian surgical subspecialty board examinations: a comparative study of ChatGPT-4o, ChatGPT-5, and Gemini 2.5 Flash. Scientific Reports. 2026;16:21139. doi:10.1038/s41598-026-51934-9.\par
\noindent\textbf{C29.} Kumarappan A, Golnari PA, Wen W, et al. DevBench: A Realistic, Developer-Informed Benchmark for Code Generation Models. arXiv. 2026. arXiv:2601.11895v3.\par
\noindent\textbf{C30.} Chen J, Dong Y, Li H, et al. Benchmarking LLM-as-a-Judge for Long-Form Output Evaluation. arXiv. 2026. arXiv:2606.01629v2.\par
\noindent\textbf{C31.} Colbran S, Jha M, Schiavone C. Understanding student perspectives on generative AI chatbots: a human-centred mixed-methods study in higher education. International Journal of Educational Technology in Higher Education. 2026;23:28. doi:10.1186/s41239-026-00605-w.\par
\noindent\textbf{C32.} Kayaci ST, Ilhan HO, Tassoker M, et al. Benchmarking GPT-5, Gemini 2.5 Pro, Grok 4, and other LLMs on pediatric dentistry questions from a dental specialization exam. Scientific Reports. 2026. doi:10.1038/s41598-026-57392-7.\par
\noindent\textbf{C33.} Van Vlasselaer M, Van Droogenbroeck F, Spruyt B. Who wrote this? Evaluating the reliability of AI detection tools in higher education. International Journal for Educational Integrity. 2026;22:16. doi:10.1007/s40979-026-00226-w.\par
\noindent\textbf{C34.} Güdül KF, Arslan B. Temporal consistency of large language model responses to restorative dentistry questions from the Turkish dental specialty examination. Scientific Reports. 2026. doi:10.1038/s41598-026-61419-4.\par
\noindent\textbf{C35.} Okonkwo D, Hodgson M, David TI, Ihejirika SA. Capabilities of Claude Fable 5 on Biomedical Challenge Problems. arXiv. 2026. arXiv:2607.10849v1.\par
\noindent\textbf{C36.} Güçlü M, Yelbay M, Karadağ İ. Performance of large language models in a high-stakes dental assessment: evidence from the Turkish dentistry specialization examination. Scientific Reports. 2026. doi:10.1038/s41598-026-62380-y.\par
\noindent\textbf{C37.} Guo Y, Wu M, Cao Y, et al. BackendForge: Benchmarking Agentic End-to-End Code Generation with Backend Services. arXiv. 2026. arXiv:2607.11042v1.\par
\noindent\textbf{C38.} Uçar S-Ş, Aldabe I, Aranberri N, De Clercq O. High-Order Question Generation in a Multilingual Educational Context. In: Proceedings of the Fifteenth Language Resources and Evaluation Conference (LREC 2026). 2026:760-769. doi:10.63317/56rihjwb6jq7; arXiv:2607.13901v1.\par
\noindent\textbf{C39.} Nogueira RP, Vieira M, Campos JR. PROBE: Benchmarking Code Generation in Large Language Models. arXiv. 2026. arXiv:2607.13820v1.\par
\noindent\textbf{C40.} Xiao J, Zhang Z, Hou H, et al. VisualRepair: Dynamic Tool Calling and Region Focusing for Visual Software Issue Repair. arXiv. 2026. arXiv:2607.14075v1.\par
\subsection{Paper-level coding}
The records were analysed as one corpus. Route and task are descriptive attributes. Model age is the estimated number of days between release of the newest tested named generation or immutable snapshot and the record's publication or posting date. For mutable product labels, the estimate may not identify the backend served at execution.

\begin{landscape}
\scriptsize
\setlength{\tabcolsep}{3pt}
\renewcommand{\arraystretch}{1.12}
\begin{longtable}{|>{\RaggedRight\arraybackslash}p{0.22\linewidth}|>{\RaggedRight\arraybackslash}p{0.14\linewidth}|>{\RaggedRight\arraybackslash}p{0.29\linewidth}|>{\RaggedRight\arraybackslash}p{0.28\linewidth}|}
\caption{Paper-level coding for the 40-record corpus.}\label{tab:corpus}\\
\hline
\textbf{Study and date} & \textbf{Route and task} & \textbf{Models and execution information} & \textbf{Estimated currency and reporting note} \\
\hline
\endfirsthead
\multicolumn{4}{c}{\textit{Table \thetable\ continued from previous page}}\\
\hline
\textbf{Study and date} & \textbf{Route and task} & \textbf{Models and execution information} & \textbf{Estimated currency and reporting note} \\
\hline
\endhead
\hline
\multicolumn{4}{r}{\textit{Continued on next page}}\\
\endfoot
\hline
\endlastfoot
C1. Bedi et al., Fidelity of Medical Reasoning\newline 8 Aug 2025 & Journal | clinical reasoning & DeepSeek-R1; o3-mini; Claude 3.5 Sonnet; Gemini 2.0 Flash; GPT-4o; Llama 3.3 70B. Runs in Mar-Apr 2025. & Newest named family about 189 days old; GPT-5 launched the previous day. Family names only; no controlled refresh found. \\ 
\hline
C2. Performance-Efficiency Optimized Routing\newline 18 Aug 2025 & Preprint | model routing & GPT-5-chat/medium; Claude Opus 4.1/Sonnet 4; Gemini 2.5 Pro/Flash; Qwen3-235B. & Newest model 11 days old. Very fresh, but OpenRouter backends and exact query dates were not pinned. \\ 
\hline
C3. Shyr et al., Rare Disease Diagnosis\newline 22 Aug 2025 & Journal | diagnosis & ChatGPT-4o observation dated 6 Aug 2024; Llama 3.1 8B Instruct observation dated 18 Dec 2024. & Newest named family about 395 days old; recorded observations were 381 and 247 days old. Azure GPT-4o deployment not identified; no rerun. \\ 
\hline
C4. Kliem et al., Status Epilepticus Guidelines\newline 28 Aug 2025 & Journal | clinical guidance & Free ChatGPT-3.5 in 2023-24; GPT-4o and GPT-5 added in Aug 2025. & GPT-5 about 21 days old. Clear late refresh, but models were unpinned and repetition counts differed. \\ 
\hline
C5. Xiao and Wang, Predicting Human Social Decisions\newline 2 Sep 2025 & Journal | social behaviour & ChatGPT-3.5, May-Nov 2023; GPT-4, Apr-Aug 2024; GPT-4o validation, 16-18 Jul 2025. & Floating GPT-4o family about 477 days old, but validation only 46 days old. Apparent late sensitivity test; mutable web products remain unreproducible. \\ 
\hline
C6. Liu et al., LLM Multiagent Order Sets\newline 23 Sep 2025 & Journal | clinical agents & GPT-4o for every agent and the judge; development and evaluation during 2024. & GPT-4o family about 498 days old; GPT-5 available. No snapshot, exact inference dates or refresh. \\ 
\hline
C7. GDPval\newline 25 Sep 2025 & Lab report | work tasks & GPT-4o, o4-mini, o3, GPT-5, Claude Opus 4.1, Gemini 2.5 Pro and Grok 4. & Newest model 49 days old. Presented as a first version; systems also differed in tools and interfaces. \\ 
\hline
C8. IOAA Gold Medal Performance\newline 6 Oct 2025 & Preprint | scientific reasoning & GPT-5, o3, Gemini 2.5 Pro, Claude Opus 4.1 and Claude Sonnet 4. & Newest model 60 days old. Sonnet 4 had been superseded seven days earlier; no update plan. \\ 
\hline
C9. Zubaer et al., GPT-4 Ranking Open-Text Answers\newline 8 Oct 2025 & Journal | assessment & GPT-4 and GPT-3.5-turbo-instruct; core runs in 2023. & GPT-4 family about 939 days old; outputs more than two years old. A later drift check was reported, but its model and date were not identified. \\ 
\hline
C10. Gholami et al., Ophthalmology MCQ Generation\newline 16 Oct 2025 & Journal | clinical assessment & Azure GPT-4o 2024-05-01-preview. & Exact build 533 days old. Strong pinning, but no current-model bridge test. \\ 
\hline
C11. Arora et al., Divergent and Convergent Thinking\newline 22 Oct 2025 & Journal | creativity & ChatGPT-4o, DeepSeek-V3 and Gemini 2.0; run in Feb 2025. & Newest family about 300 days old at publication. Clear data-to-publication lag; no snapshot IDs or refresh. \\ 
\hline
C12. Idan et al., Complex Medical Queries\newline 29 Oct 2025 & Journal | medical queries & GPT-4o, Gemini 2.0 and Claude 3.5 Sonnet through public web interfaces. & Newest family about 322 days old. No query dates, snapshots or refresh. \\ 
\hline
C13. Two-Faced Social Agents\newline 19 Nov 2025 & Preprint | social agents & GPT-5, Claude Sonnet 4.5 and Gemini 2.5 Flash. & Newest model 51 days old. GPT-5.1 and Gemini 3 appeared immediately before posting; no rerun. \\ 
\hline
C14. Skjervold et al., DiaGuide-LLM\newline 28 Nov 2025 & Journal | diagnosis support & Azure GPT-4o 2024-08-06; responses generated Sep 2024. & Exact snapshot 479 days old. Explicit point-in-time framing and future-update proposal, but no completed rerun. \\ 
\hline
C15. Sheikhalishahi et al., Persian Internal-Medicine Exams\newline 3 Dec 2025 & Journal | medical exams & GPT-4o, GPT-5 and Gemini 2.5 Flash; run in Sep 2025. & GPT-5 about 118 days old; GPT-5.1 had appeared. Relatively current, but snapshots unidentified and no refresh reported. \\ 
\hline
C16. Reasoning Models Ace the CFA Exams\newline 9 Dec 2025 & Preprint | financial reasoning & GPT-3.5/4/4o/5; Gemini 2.5/3 Pro; Grok 4; Claude Opus 4.1; DeepSeek V3.1; o4-mini judge. & Newest model about 21 days old. The attempted snapshot table contains several identifiers or dates that do not align cleanly with provider records. \\ 
\hline
C17. Are We on the Right Way to Assessing LLM-as-a-Judge?\newline 17 Dec 2025 & Preprint | model judging & Thirteen judges across GPT, Gemini, Claude, Qwen, DeepSeek and Llama families. & Newest model about 118 days old. Older models were partly intentional capability tiers; no query dates or refresh. \\ 
\hline
C18. Bai et al., GPTutor\newline 19 Dec 2025 & Journal | tutoring & Azure gpt-4o-mini-2024-07-18. & Exact snapshot about 519 days old. No successor retest; later correction was administrative. \\ 
\hline
C19. Patel et al., Successive GPT Models in Medical Cases\newline 6 Jan 2026 & Journal | medical cases & Dated GPT-3.5, GPT-4 Turbo, GPT-4o, GPT-4.1, o3 and GPT-5 identifiers, including gpt-5-2025-08-07. & GPT-5 snapshot about 152 days old; GPT-5.2 already available. Clear refresh, strong reporting and explicit lifecycle-monitoring argument. \\ 
\hline
C20. ChipBench\newline 29 Jan 2026 & Preprint | chip design & GPT-3.5/4o/5/5.2; Claude 4.5 variants; Gemini 2.5/3; DeepSeek; Llama, with several exact OpenAI IDs. & Newest model 49 days old. Unusually strong release-date and API reporting; no later-model refresh. \\ 
\hline
C21. Sajja et al., AI Learning Assistants in Engineering\newline 6 Feb 2026 & Journal | learning assistant & GPT-4o, text-embedding-3-large and Nougat parsing. & GPT-4o family about 634 days old. No semester year, snapshot or model refresh. \\ 
\hline
C22. Cheung and Zhang, Nature of Science in GPT-4o\newline 11 Feb 2026 & Journal | science education & Subscription ChatGPT/GPT-4o, queried 25 Oct 2024. & Family about 639 days old; data-to-publication lag about 15.6 months. Revised through Nov 2025 without rerunning the model. \\ 
\hline
C23. Oh et al., Classroom AI\newline 3 Mar 2026 & Journal | classroom support & GPT-4o mini/4o; Llama 3.1/3.3; Qwen 2.5; Gemma 2; Phi-4; Mixtral; GPT-2 XL; embedding model. & Newest principal family about 446 days old. No snapshots, run dates or bridge to GPT-5-era systems. \\ 
\hline
C24. Long Reasoning Chains and Factuality Judgment\newline 8 Apr 2026 & Preprint | factuality judging & Qwen3; DeepSeek V3.1; Llama 3.1; GLM4/Z1; GPT-4o; Claude Sonnet 4.5; Qwen2.5 verifier. & Newest model about 191 days old. GPT-4o and Sonnet 4.5 were described in a strong tier after later generations existed. \\ 
\hline
C25. Rao et al., Clinical Reasoning Tasks\newline 13 Apr 2026 & Journal | clinical reasoning & Twenty-one systems across GPT, Claude, DeepSeek, Gemini and Grok families. & Newest families about 145 days old. Broad coverage but mostly product names; future-facing framework without a completed refresh. \\ 
\hline
C26. Saez et al., AI-Assisted Grading\newline 18 Apr 2026 & Journal | grading & gpt-4o-2024-08-06; gpt-4o-mini-2024-07-18; Llama 3.1/3.3; DeepSeek-R1/V3. & Newest family about 453 days old. Exact OpenAI snapshots and reusable framework, but no current-model bridge. \\ 
\hline
C27. Zhang et al., Liver-Cirrhosis Benchmark\newline 22 Apr 2026 & Journal | clinical benchmark & Claude Opus 4, DeepSeek-R1, Gemini 2.5 Pro, GPT-5, o3 and Grok 4; experiments Jul-Sep 2025. & Newest model about 258 days old; experimental evidence roughly 7-10 months old. Recent publication, historical panel. \\ 
\hline
C28. Sheikhalishahi et al., Persian Surgical Board Exams\newline 8 May 2026 & Journal | surgical exams & GPT-4o, GPT-5 and Gemini 2.5 Flash; all runs in Sep 2025. & GPT-5 about 274 days old; about eight months from testing to publication. No dated snapshot or bridge test. \\ 
\hline
C29. DevBench, version 3\newline 16 May 2026 & Preprint | software development & Updated panel including GPT-5.5/5.4 variants, Claude Opus 4.7/Sonnet 4.6, DeepSeek V4 Pro, Llama 4, Mistral 3.5 and Qwen3.6. & Newest model about 22 days old. Rare completed refresh, although benchmark construction also changed between versions. \\ 
\hline
C30. LongJudgeBench\newline 1 Jun 2026 & Preprint | long-context judging & Qwen3, GPT-4o mini, GPT-5.2, DeepSeek V4 Flash, GLM-5.1 and Kimi-K2.6. & Newest model about 38 days old. Current by normal standards, but GPT-5.2 had already been superseded by GPT-5.5. \\ 
\hline
C31. Colbran et al., Student Perspectives on Chatbots\newline 10 Jun 2026 & Journal | student experience & GPT-4o-powered Jordan Chatbot. GPT-5 was used only to polish manuscript language. & Intervention family about 758 days old. The newer model edited the paper but was not used for a bridge test. \\ 
\hline
C32. Kayaci et al., Paediatric-Dentistry Benchmark\newline 11 Jun 2026 & Journal | dental benchmark & Eleven models including GPT-5, Gemini 2.5 Pro, Grok 4 and DeepSeek-R1. & Newest model about 308 days old. Fable 5 had appeared only two days earlier, making that particular omission unavoidable. \\ 
\hline
C33. Van Vlasselaer et al., AI Detection Tools\newline 29 Jun 2026 & Journal | AI detection & Paper-reported GPT-4o Deep Research and GPT-4o/consumer ChatGPT; outputs generated May 2025. & Floating newest family about 777 days old; outputs 13 months old. The paper acknowledges obsolescence and names later models for future work, but does not rerun. \\ 
\hline
C34. Güdül and Arslan, Restorative-Dentistry Consistency\newline 7 Jul 2026 & Journal | dental consistency & GPT-5.1, Gemini 2.5 Pro, Copilot and DeepSeek-v3.2. & Newest family about 237 days old. GPT-5.5 existed; GPT-5.6 arrived two days after publication. \\ 
\hline
C35. Okonkwo et al., Fable 5 Biomedical Challenge Problems\newline 12 Jul 2026 & Preprint | biomedical benchmark & Claude Fable 5, two Claude predecessors and GPT-5. & Fable 5 about 33 days from original launch and 11 days from redeployment. Strong example of fast preprinting; routing and refusal outcomes are part of the tested system. \\ 
\hline
C36. Güçlü et al., High-Stakes Dental Assessment\newline 13 Jul 2026 & Journal | dental assessment & GPT-5, Gemini 2.5 Pro, Claude Sonnet 4.5 and DeepSeek-v3.2. & Newest panel roughly 287 days old by the coding rule. Accepted only three days earlier; a frontier note or sentinel bridge was more realistic than a full rerun. \\ 
\hline
C37. BackendForge\newline 13 Jul 2026 & Preprint | backend coding & GPT-5.5 was the strongest tested system. & Newest model about 80 days old; GPT-5.6 had appeared four days earlier. Even a fresh preprint can be overtaken during final preparation. \\ 
\hline
C38. Uçar, High-Order Question Generation\newline 15 Jul 2026 & Preprint | question generation & Llama 3.1 and GPT-4; no exact IDs or run date. & Newest family about 722 days old. A new posting is not automatically a current model evaluation. \\ 
\hline
C39. PROBE\newline 15 Jul 2026 & Preprint | benchmark framework & Newest named system GPT-4.1-mini, with older comparison models. & Newest family about 457 days old. Framework designed to evolve, but the published panel was already historical. \\ 
\hline
C40. VisualRepair\newline 15 Jul 2026 & Preprint | visual software repair & Exact o3-2025-04-16 and dated comparison snapshots. & Newest snapshot about 455 days old. Strong pinning and reproducibility, but an old model backbone at posting. \\ 
\hline
\end{longtable}
\noindent\textbf{Coding note.} Counts and age estimates apply only to this purposive corpus. Release dates were drawn from public provider records; model age and supersession follow the definitions in Section 2. Calendar age is not a direct measure of task capability or claim validity.
\end{landscape}

\end{document}